\documentclass[twoside,11pt]{article}

\usepackage[ruled,vlined]{algorithm2e}
\usepackage{mathtools}
\mathtoolsset{showonlyrefs=true}

\usepackage{jmlr2e}

\newcommand{\argmin}{\operatorname*{argmin}}

\newcommand{\E}{\mathbb{E}}

\newcommand{\s}{\mathcal{S}}
\newcommand{\A}{\mathcal{A}}
\newcommand{\p}{P}

\newcommand{\rr}{\mathcal{R}}

\ShortHeadings{Anderson Acceleration for Reinforcement Learning}{Geist and Scherrer}
\firstpageno{1}

\begin{document}

\title{Anderson Acceleration for Reinforcement Learning}

\author{\name Matthieu Geist \email matthieu.geist@univ-lorraine.fr \\
       \addr Universit\'e de Lorraine, CNRS, LIEC, F-57000 Metz, France\\
       (Now at Google Brain)\\
\AND
\name Bruno Scherrer \email bruno.scherrer@inria.fr \\
	\addr Universit\'e de Lorraine, CNRS, Inria, IECL, F-54000 Nancy, France}

\ewrlheading{14}{2018}{October 2018, Lille, France}{Matthieu Geist and Bruno Scherrer}

\maketitle

\begin{abstract}
\citet{anderson1965iterative} acceleration is an old and simple method for accelerating the computation of a fixed point. However, as far as we know and quite surprisingly, it has never been applied to dynamic programming or reinforcement learning. In this paper, we explain briefly what Anderson acceleration is and how it can be applied to value iteration, this being supported by preliminary experiments showing a significant speed up of convergence, that we critically discuss. We also discuss how this idea could be applied more generally to (deep) reinforcement learning. 
\end{abstract}


\section{Introduction}

Reinforcement learning (RL)~\citep{sutton1998reinforcement} is intrinsically linked to fixed-point computation: the optimal value function is the fixed point of the (nonlinear) Bellman optimality operator, and the value function of a given policy is the fixed point of the related (linear) Bellman evaluation operator. Most of the time, these fixed points are computed recursively, by applying repeatedly the operator of interest. Notable exceptions are the evaluation step of policy iteration and the least-squares temporal differences (LSTD) algorithm\footnote{In the realm of deep RL, as far as we know, all fixed points are computed iteratively, there is no LSTD.}~\citep{Bradtke:1996}.

\citet{anderson1965iterative} acceleration (also known as Anderson mixing, Pulay mixing, direct inversion on the iterative subspace or DIIS, among others\footnote{Anderson acceleration and variations have been rediscovered a number of times in various communities in the last 50 years. \citet{walker2011anderson} provide a brief overview of these methods, and a close approach has been recently proposed in the machine learning community~\citep{scieur2016regularized}.}) is a method that allows speeding up the computation of such fixed points. The classic fixed-point iteration applies repeatdly the operator to the last estimate. Anderson acceleration considers the $m$ previous estimates. Then, it searches for the point that has minimal residual within the subspace spanned by these estimates, and applies the operator to it. This approach has been successfully applied to fields such as electronic structure computation or computational chemistry, but it has never been applied to dynamic programming or reinforcement learning, as far as we know. For more about Anderson acceleration, refer to~\citet{walker2011anderson}, for example.

\section{Anderson Acceleration for Value Iteration}

In this section, we briefly review Markov decision processes, value iteration, and show how Anderson acceleration can be used to speed up convergence.

\subsection{Value Iteration}

Let $\Delta_X$ be the set of probability distributions over a finite set $X$ and $Y^X$ the set of applications from $X$ to the set $Y$. By convention, all vectors are column vectors. A Markov Decision Process (MDP) is a tuple $\{\s,\A,\p,\rr, \gamma\}$, where $\s$ is the finite state space, $\A$ is the finite action space, $\p\in(\Delta_\s)^{\s\times \A}$ is the Markovian transition kernel ($\p(s'|s,a)$ denotes the probability of transiting to $s'$ when action $a$ is applied in state $s$), $\rr\in\mathbb{R}^{\s\times\A}$ is the bounded reward function ($\rr(s,a)$ represents the local benefit of doing action $a$ in state $s$) and $\gamma\in(0,1)$ is the discount factor.

A stochastic policy $\pi \in (\Delta_\A)^\s$ associates a distribution over actions to each state (deterministic policies being a special case of this). The policy-induced reward and transition kernels, $\rr_\pi\in\mathbb{R}^\s$ and $\p_\pi \in (\Delta_\s)^\s$, are defined as
\begin{equation}
\rr_\pi(s) = \E_{\pi(.|s)}[\rr(s,A)]
\text{ and } \p_\pi(s'|s) = \E_{\pi(.|s)}[\p(s'|s,A)].
\end{equation}
The quality of a policy is quantified by the associated value function $v_\pi\in\mathbb{R}^\s$:
\begin{equation}
v_\pi(s) = \E[\sum_{t\geq 0} \gamma^t \rr_\pi(S_t)|S_0 = s, S_{t+1}\sim \p_\pi(.|S_t)].
\end{equation}
The value $v_\pi$ is the unique fixed point of the Bellman operator $T_\pi$, defined as $T_\pi v = \rr_\pi + \gamma \p_\pi v$ for any $v\in\mathbb{R}^\s$.

Let define the second Bellman operator $T$ as, for any $v\in\mathbb{R}^\s$, $Tv = \max_{\pi\in(\Delta_\A)^\s} T_\pi v$. This operator is a $\gamma$-contraction (in supremum norm), so the iteration
\begin{equation}
v_{k+1} = T v_k
\end{equation}
converges to its unique fixed-point $v_* = \max_\pi v_\pi$, for any $v_0\in\mathbb{R}^{\s}$. This is the value iteration algorithm.

\subsection{Accelerated Value Iteration}

Anderson acceleration is a method that aims at accelerating the computation of the fixed point of any operator. Here, we describe it  considering the Bellman operator $T$, which provides an accelerated value iteration algorithm.

Assume that estimates have been computed up to iteration $k$, and that in addition to $v_k$ the $m$ previous estimates $v_{k-1}, \dots, v_{k-m}$ are known. The coefficient vector $\alpha^{k+1}\in\mathbb{R}^{m+1}$ is defined as follows:
\begin{equation}
\alpha^{k+1} = \argmin_{\alpha \in \mathbb{R}^{m+1}} \left\|\sum_{i=0}^m \alpha_i (Tv_{k-m+i} - v_{k-m+i})\right\| \text{ s.t. } \sum_{i=0}^m \alpha_i = 1.
\end{equation}
Notice that we don't impose a positivity condition on the coefficients.
We will consider practically the $\ell_2$-norm for this problem, but it could be a different norm (for example $\ell_1$ or $\ell_\infty$, in which case the optimization problem is a linear program).
Then, the new estimate is given by:
\begin{equation}
v_{k+1} = \sum_{i=0}^m \alpha_i^{k+1} T v_{k-m+i}.
\end{equation}
The resulting Anderson accelerated value iteration is summarized in Alg.~\ref{alg}. Notice that the solution to the optimization problem can be obtained analytically for the $\ell_2$-norm, using the Karush-Kuhn-Tucker conditions. With the notations of Alg.~\ref{alg} and writting $\mathbf{1}\in\mathbb{R}^{m_k+1}$ the vector with all components equal to one, it is
\begin{equation}
\alpha^{k+1} = \frac{(\Delta_k^\top \Delta_k)^{-1}\mathbf{1}}{\mathbf{1}^\top (\Delta_k^\top \Delta_k)^{-1}\mathbf{1}}.\label{eq:alpha}
\end{equation}
This can be regularized to avoid ill-conditioning.

\begin{algorithm}[tbh]
	\SetKwInOut{Input}{given}
	\Input{$v_0$ and $m\geq 1$}
	Compute $v_1 = T v_0$\;
	\For{$k=1,2,3\dots$}{
		Set $m_k = \min(m,k)$\;
		Set $\Delta_k = [\delta_{k-m_k},\dots, \delta_k]\in\mathbb{R}^{\s\times (m+1)}$ with $\delta_i = T v_i - v_i\in\mathbb{R}^\s$\;
		Solve $\min_{\alpha\in\mathbb{R}^{m+1}} \|\Delta_k \alpha\|$ s.t. $\sum_{i=0}^{m_k} \alpha_k = 1$\;
		Set $v_{k+1} = \sum_{i=0}^{m_k} \alpha_i T v_{k-m_k+i}$\;
	}
	\caption{Anderson Accelerated Value Iteration\label{alg}}
\end{algorithm}

The rationale of this acceleration scheme is better understood with an affine operator. We consider here the Bellman evaluation operator $T_\pi$. Given the current and the $m$ previous estimates, define
\begin{equation}
\tilde{v}^\alpha_{k+1} = \sum_{i=0}^m \alpha_i v_{k-m+i} \text{ with } \sum_{i=0}^m \alpha_i = 1.
\end{equation}
Thanks to this constraint, for an affine operator (here $T_\pi$), we have that
\begin{equation}
T_\pi \tilde{v}^\alpha_{k+1}  = \sum_{i=0}^m \alpha_i T_\pi v_{k-m+i}.
\end{equation}
Then, one searches for a vector $\alpha$ (satisfying the constraint) that minimizes the residual
\begin{equation}
\|T_\pi \tilde{v}^\alpha_{k+1} - \tilde{v}^\alpha_{k+1}\| = \|\sum_{i=0}^m \alpha_i (T_\pi v_{k-m+i} - v_{k-m+i})\|. 
\end{equation} 
Eventually, the new estimate is obtained by applying the operator to the vector $\tilde{v}^\alpha_{k+1}$ of minimal residual.

The same approach can be applied (heuristically) to non-affine operators. The convergence of this scheme has been studied (\textit{e.g.},~\citet{toth2015convergence}) and it can be linked to quasi-Newton methods~\citep{fang2009two}.

\subsection{Preliminary Experimental Results}

We consider Garnet problems~\citep{archibald1995generation,bhatnagar2009natural}. They are a class of randomly built MDPs meant to be totally abstract while remaining representative of the problems that might be encountered in practice. Here, a Garnet $G(|\s|, |\A|, b)$ is specified by the number of states, the number of actions and the branching factor. For each $(s,a)$ couple, $b$ different next states are chosen randomly and the associated probabilities are set by randomly partitioning the unit interval. The reward is null, except for $10\%$ of states where it is set to a random value, uniform in $(1,2)$.

We generate 100 random MDPs $G(100,4,3)$ and set $\gamma$ to $0.99$. For each MDP, we apply value iteration (denoted as $m=0$ in the graphics) and Anderson accelerated value iteration for $m$ ranging from 1 to 9. The inital value function $v_0$ is always the null vector. We run all algorithms for $250$ iterations, and measure the normalised error for algorithm \texttt{alg} at iteration $k$, $\frac{\|v_* - v^{\text{\texttt{alg}}}_k\|_1}{\|v_*\|_1}$,
where $v_*$ stands for the optimal value function of the considered MDP.

\begin{figure}[tbh]
	\begin{minipage}[l]{0.32\linewidth}
		\includegraphics[width=\linewidth]{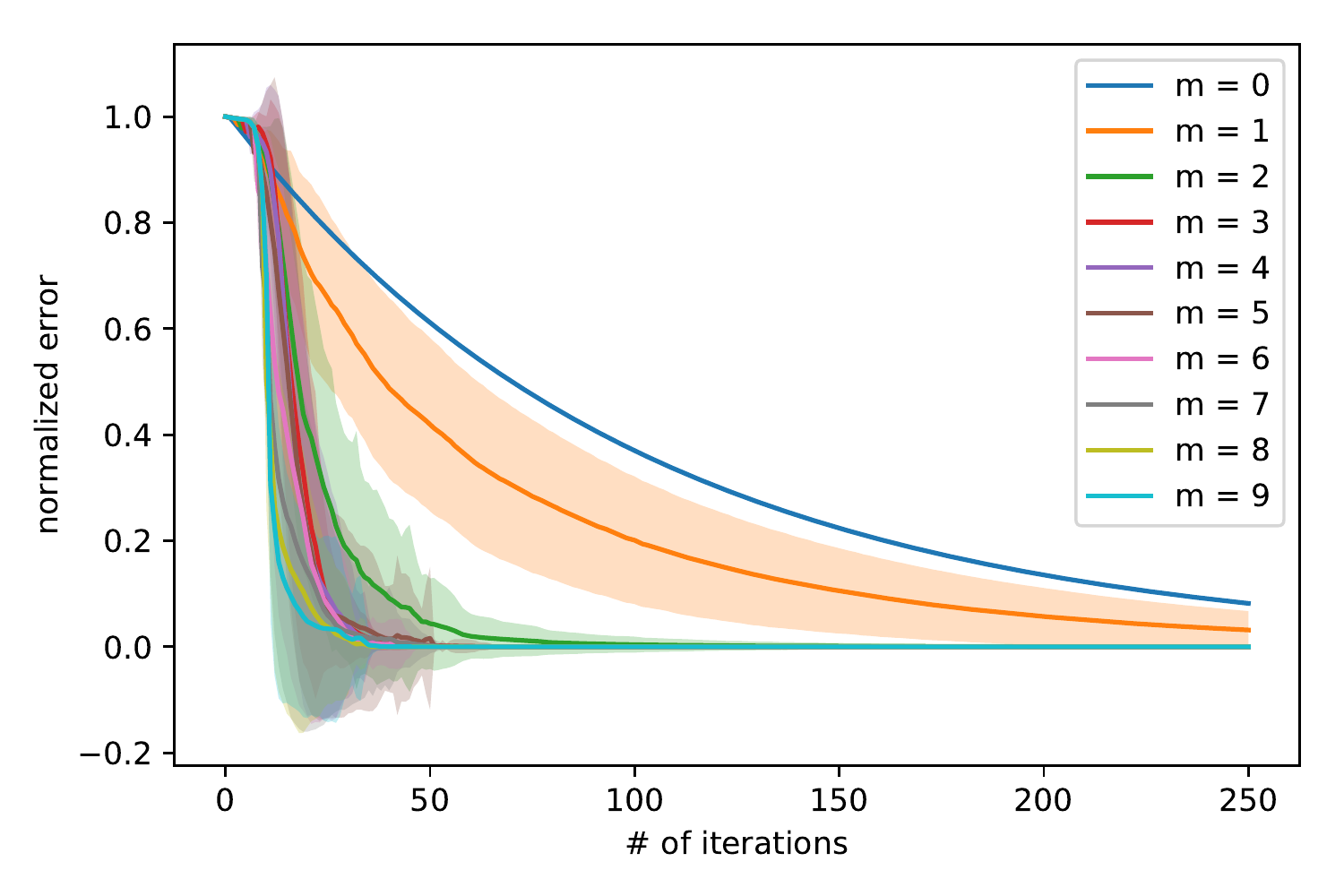}
		\center{a. Normalized error\\~.}
	\end{minipage}
	\begin{minipage}[l]{0.32\linewidth}
		\includegraphics[width=\linewidth]{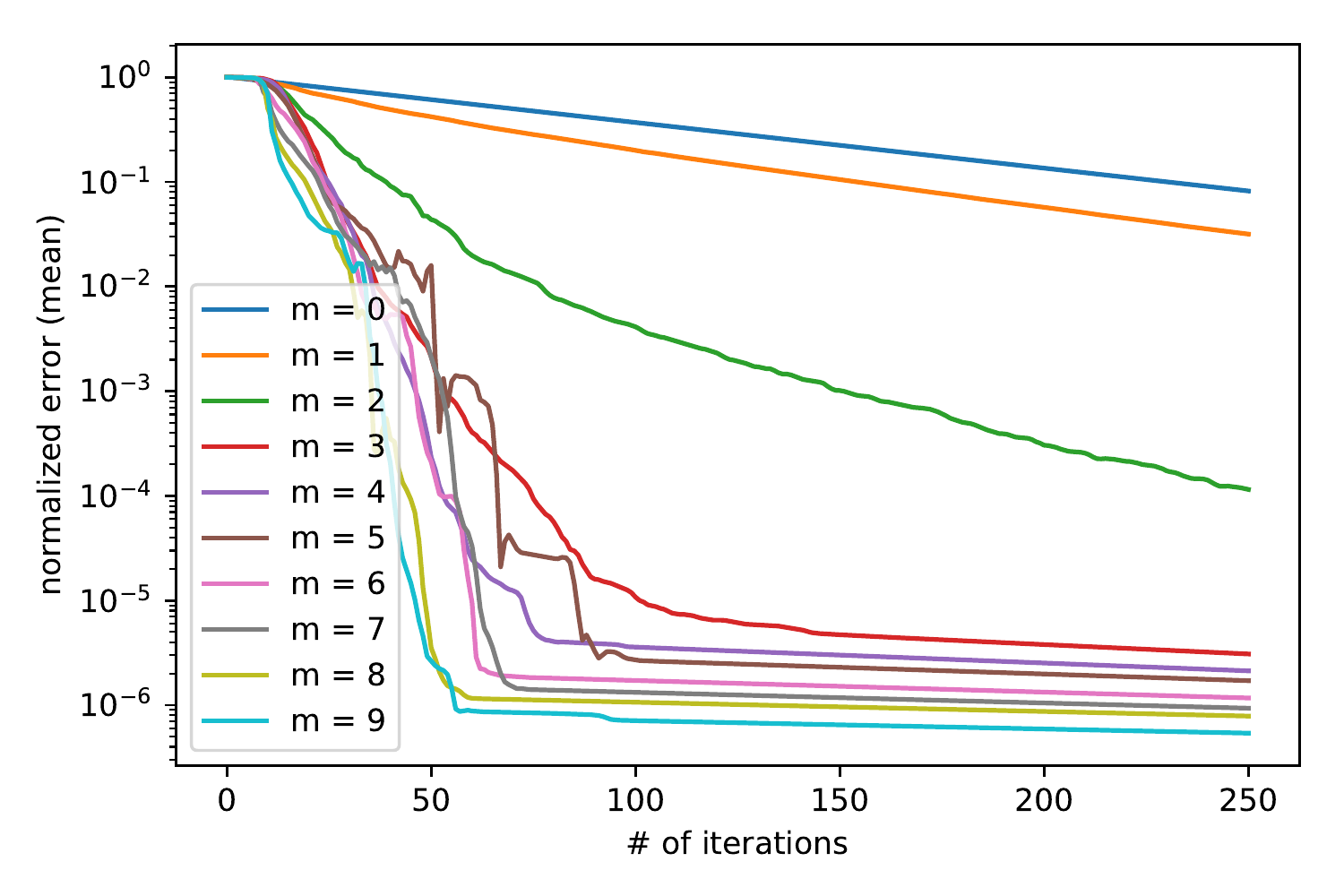}
		\center{b. Normalized error\\~~~~(mean, log-scale).}
	\end{minipage}
	\begin{minipage}[l]{0.32\linewidth}
		\includegraphics[width=\linewidth]{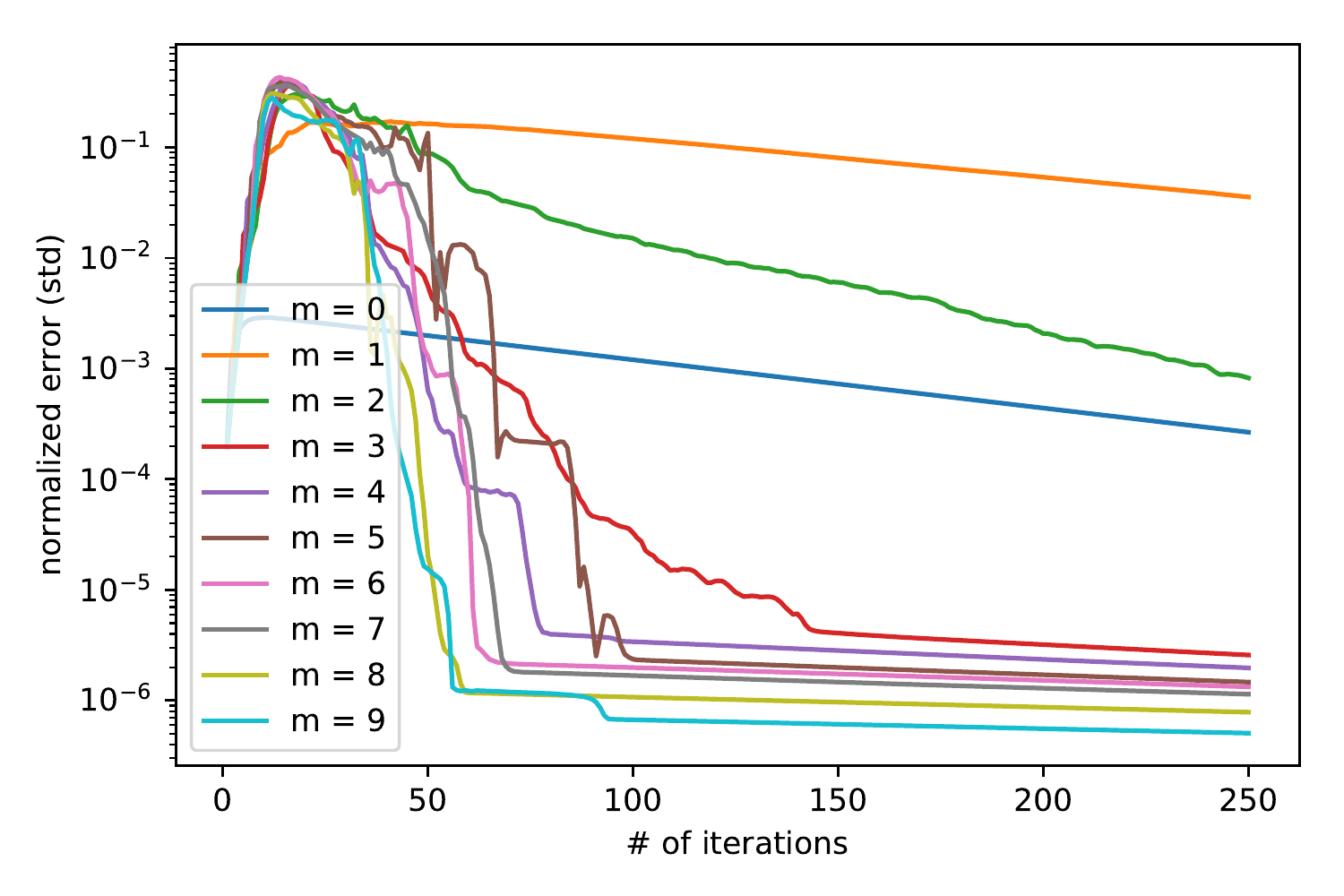}
		\center{c. Normalized error\\(std, log-scale).}
	\end{minipage}
	\caption{Results on the Garnet problems. \label{fig}}
\end{figure}

Fig.~\ref{fig} shows the results. Fig.~\ref{fig}.a shows how the normalized error evolves  with the number of iterations (recall that $m=0$ stands for classic value iteration). Shaded areas correspond to standard deviations and lines to means (due to randomness of the MDPs, the algorithms being deterministic given the fixed initial value function). Fig.~\ref{alg}.b and~\ref{alg}.c show respectively the mean and the standard deviation of these errors, in a logarithmic scale.
One can observe that Anderson acceleration consistently offers a significant speed-up compared to value iteration, and that rather small values of $m$ ($m\approx 5$) seem to be enough.

\subsection{Nuancing the Acceleration}

We must highlight that the optimal policy is the object of interest, the value function being only a proxy to it. Regarding the value function, its level is not that important, but its relative differences are. This is addressed by the relative value iteration algorithm~\citep[Ch.~6.6]{Puterman:1994}. For a given state $s_0$, it iterates as $v_{k+\frac{1}{2}} = T v_k$, $v_{k+1} = v_{k+\frac{1}{2}} - v_{k+\frac{1}{2}}(s_0)\mathbf{1}$. It usually converges much faster than value iteration (towards $v_* - v_*(s_0)\mathbf{1}$), but the greedy policies resp. to each iterate's estimated values are the same for both algorithms. This scheme can also be easily accelerated with Anderson's approach.

\begin{figure}[tbh]
	\begin{minipage}[l]{0.32\linewidth}
		\includegraphics[width=\linewidth]{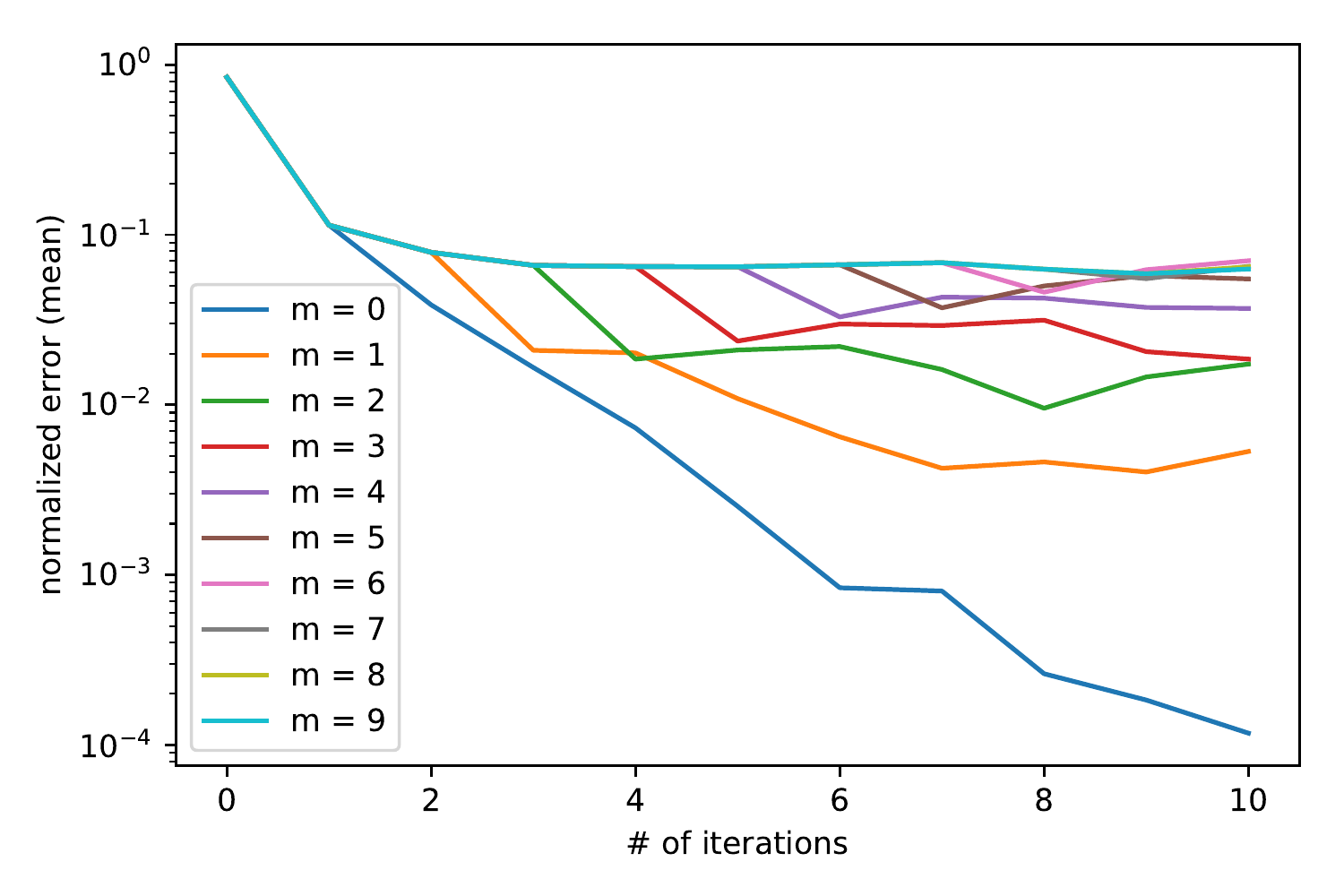}
		\center{a. Error on greedy policies (accelerated VI).}
	\end{minipage}
	\begin{minipage}[l]{0.32\linewidth}
		\includegraphics[width=\linewidth]{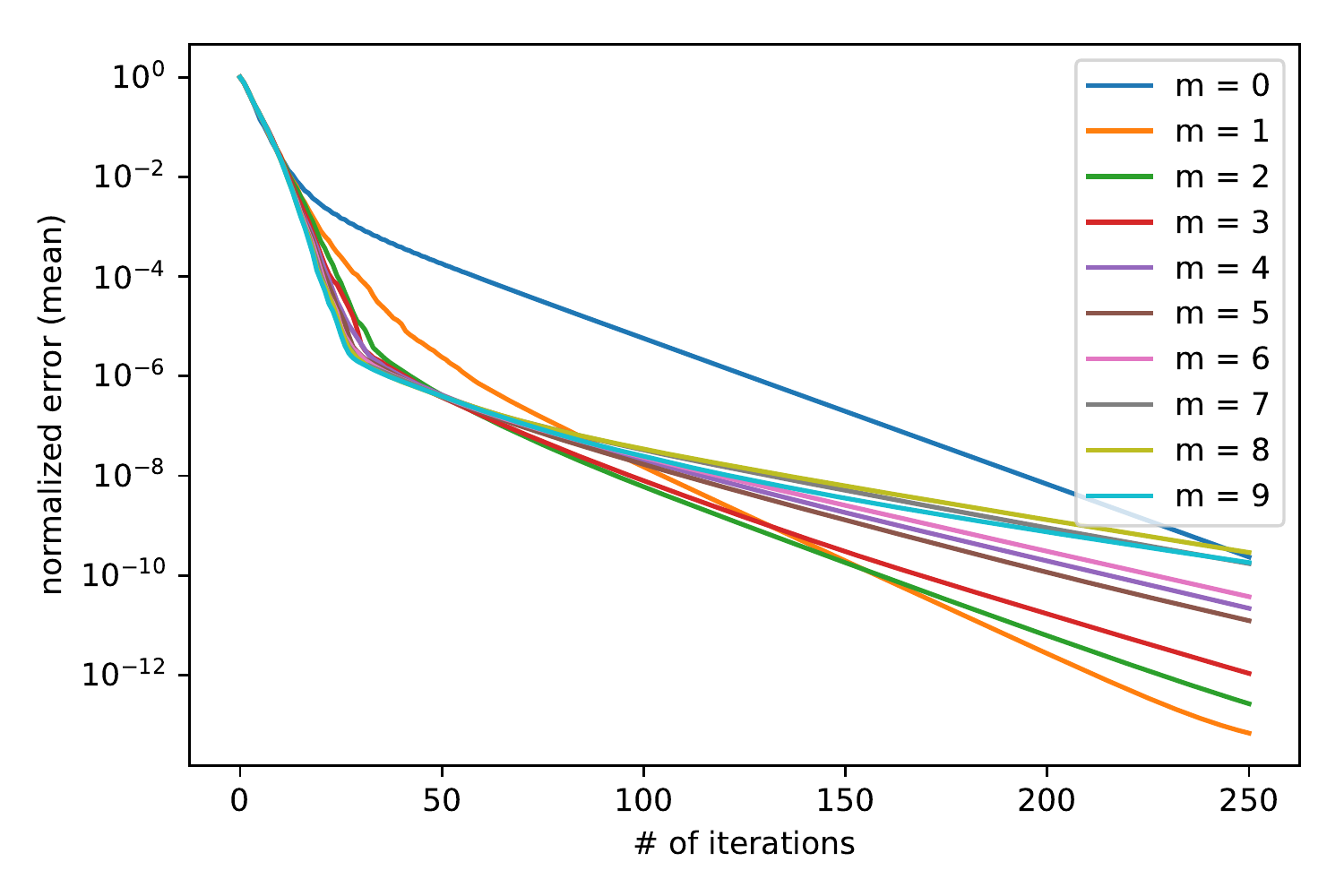}
		\center{b. Normalized error (accelerated relative VI).}
	\end{minipage}
	\begin{minipage}[l]{0.32\linewidth}
		\includegraphics[width=\linewidth]{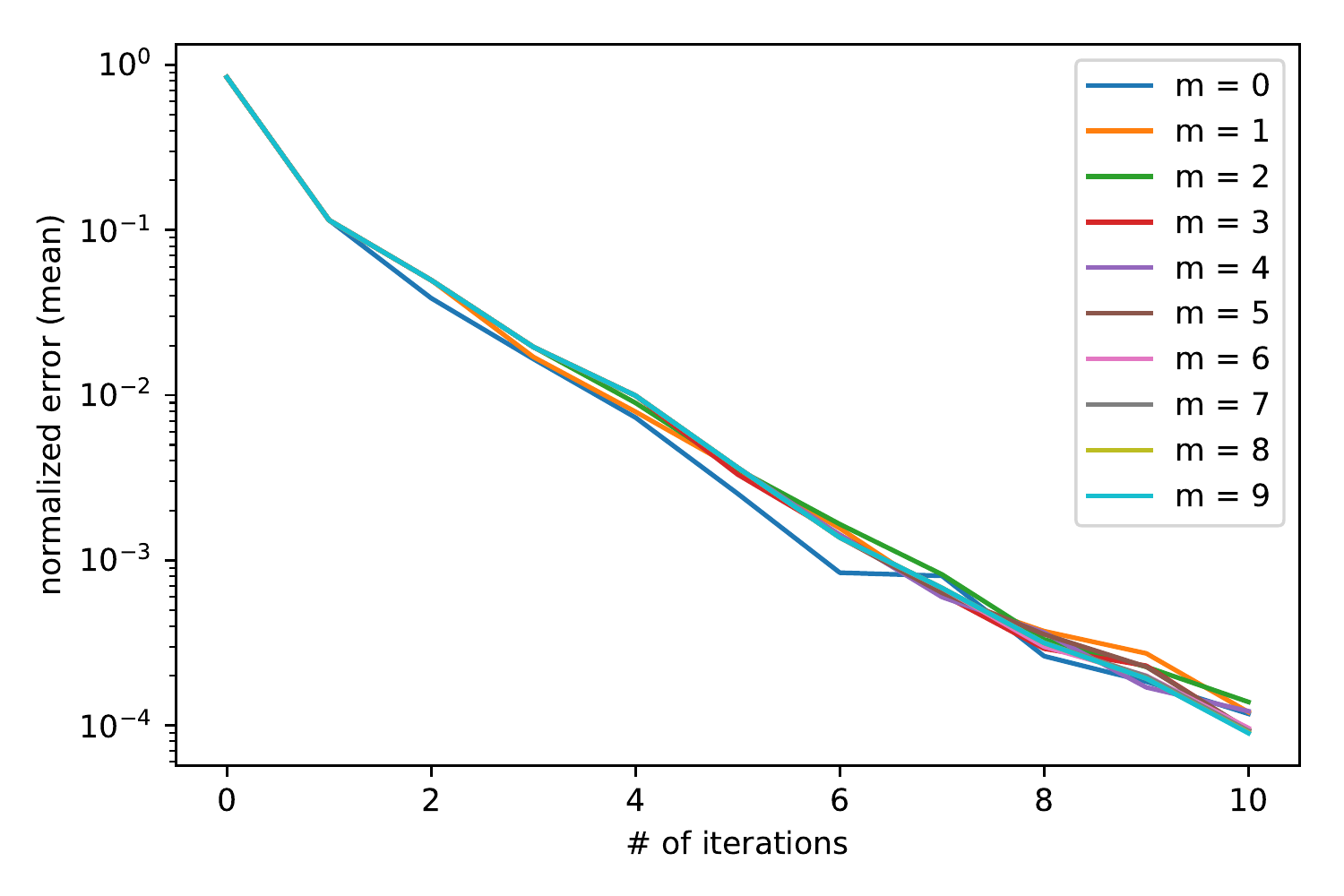}
		\center{c. Error on greedy policies (accelerated relative VI).}
	\end{minipage}
	\caption{Additional results. \label{fig_add}}
\end{figure}

We provide additional results on Fig.~\ref{fig_add} (for the same MDPs as previously). Fig.~\ref{fig_add}.a shows the error of the greedy policy, that is ${\|v_* - v_{\pi^{\text{\texttt{alg}}}_k}\|_1}/{\|v_*\|_1}$, with $\pi^{\text{\texttt{alg}}}_k$ being greedy respectively to $v^{\text{\texttt{alg}}}_k$, for the first 10 iterations (same data as for Fig.~\ref{fig}). This is what we're really interested in. One can observe that value iteration provides more quickly better solutions than Anderson acceleration. This is due to the fact that if the level of the value function converges slowly, its relative differences converge more quickly.

So, we compare relative value iteration and its accelerated counterpart in Fig.~\ref{fig_add}.b (normalized error of the estimate, not of the greedy policy), to be compared to Fig.~\ref{fig}.b. There is still an acceleration with Anderson, at least at the beginning, but the speed-up is much less than in Fig.~\ref{fig}. We compare the error on greedy policies for the same setting in Fig.~\ref{fig_add}.c, and all approaches perform equally well.

\section{Anderson Acceleration for Reinforcement Learning}
\label{sec:aa_rl}

So, the advantage of Anderson acceleration applied to exact value iteration on simple Garnet problems is not that clear. Yet, it could still be interesting for policy evaluation or in the approximate setting. We discuss briefly its possible applications to (deep) RL.

\subsection{Approximate Dynamic Programming}

Anderson acceleration could be applied to approximate dynamic programming and related methods. For example, the well-known DQN algorithm~\citep{mnih2015human} is nothing else than a (very smart) approximate value iteration approach. A state-action value function $Q$ is estimated (rather than a value function), and this function is represented as a neural network. A target network $Q_k$ is maintained, and the Q-function is estimated by solving the least-squares problem (for the memory buffer $\{(s_i,a_i,r_i,s'_i)_{1\leq i \leq n}\}$)
\begin{equation}
\frac{1}{n}\sum_{i=1}^n (y_i - Q_\theta(s_i,a_i))^2 \text{ with } y_i = r_i + \gamma \max_{a\in\A} Q_k(s'_i,a).
\end{equation} 
Anderson acceleration can be applied directly as follows. Assume that the $m+1$ previous target networks $Q_k,\dots,Q_{k-m}$ are maintained. Define for $k-m\leq j\leq k$
\begin{equation}
\delta_j = [r_1 + \gamma \max_a Q_j(s'_1,a) - Q_j(s_1,a_1), \dots, r_n + \gamma \max_a Q_j(s'_n,a) - Q_j(s_n,a_n)]^\top\in\mathbb{R}^n
\end{equation}
and $\Delta_k = [\delta_{k-m},\dots, \delta_k]\in\mathbb{R}^{n\times (m+1)}$. Solve $\alpha_{k+1}$ as in Eq.~\eqref{eq:alpha} and define for all $1\leq i\leq n$
\begin{equation}
y_i = \sum_{j=0}^n \alpha_j (r_i + \gamma \max_{a\in\A} Q_{k-m+j}(s'_i,a)).
\end{equation}
So, Anderson acceleration would modify the targets in the regression problem, the necessary coefficients being obtained with a cheap least-squares (given $m$ is small enough, as suggested by our preliminary experiments). Notice that the estimate $\alpha^{k+1}$ is biased, as being the solution to a residual problem with sampled transitions. However, if a problem, this could probably be handled with instrumental variables, giving an LSTD-like algorithm~\citep{Bradtke:1996}. Variations of this general scheme could also be envisionned, for example by computing the $\alpha$ vector on a subset of the memory replay or even on the current mini-batch, or by considering variations of Anderson acceleration such as the one of~\citet{henderson2018damped}.

This acceleration scheme could be more generally applied to approximate modified policy iteration, or AMPI~\citep{scherrer2015approximate}, that generalizes both approximate policy and value iterations. Modified policy iteration is similar to policy iteration, except that instead of computing the fixed point in the evaluation step, the Bellman evaluation operator is applied $p$ times ($p=1$ gives value iteration, $p=\infty$ policy iteration), the improvement step (computing the greedy policy) being the same (up to possible approximation). In the approximate setting, the evaluation step is usually performed by performing the regression of $p$-step returns, but it could be done by applying repeatedly the evaluation operator, this being combined with Anderson acceleration (much like DQN, but with $T_\pi$ instead of $T$).

\subsection{Policy Optimization}

Another popular approach in reinforcement learning is policy optimization, or direct policy search~\citep{deisenroth2013survey}, that maximizes $J(w) = \E_{S\sim \mu}[v_{\pi_w}(S)]$ (or a proxy), for a user-defined state distribution $\mu$, over a class of parameterized policies. This is classically done by performing a gradient ascent:
\begin{equation}
w_{k+1} = w_k + \eta \nabla_{w} J(w)|_{w=w_k}.
\label{eq:grad_ascent}
\end{equation}

This gradient is given by
$\nabla_w J(w) = \E_{S\sim d_{\mu,\pi_w},A\sim \pi}[Q_{\pi_w}(S,A)\nabla_w \ln \pi_w (A|S)]$.
Thus, it depends on the state-action value function of the current policy. This gradient can be estimated with rollouts, but it is quite common to estimate the Q-function itself. Related approaches are known as actor-critic methods (the actor being the policy, and the critic the Q-function). It is quite common to estimate the critic using a SARSA-like approach, especially in deep RL. In other words, the critic is estimated by applying repeatedly the Bellman evaluation operator. Therefore, Anderson acceleration could be applied, in the same spirit as what we described for DQN.

Yet, Anderson acceleration could also be used to speed up the convergence of the policy. Consider the gradient ascent in Eq.~\eqref{eq:grad_ascent}. This can be seen as a fixed-point iteration to solve $w = w + \eta \nabla_w J(w)$.
Anderson acceleration could thus be used to speed it up. Seeing gradient descent as a fixed point is not new~\citep{jung2017fixed}, nor is applying Anderson acceleration to speed it up~\citep{scieur2016regularized,xie2018interpolatron}. Yet, it has never been applied to policy optimization, as far as we know.

\newpage
\bibliography{anderson_rl}

\end{document}